\documentclass[11pt]{article}

\usepackage[final]{acl}

\usepackage{times}
\usepackage{latexsym}
\usepackage[T1]{fontenc}
\usepackage[utf8]{inputenc}
\usepackage{microtype}

\usepackage{amsmath}
\usepackage{hyperref}
\usepackage{url}
\usepackage{amssymb} 
\usepackage{graphicx}
\usepackage{xspace} 
\usepackage{longtable}

\usepackage{listings}
\usepackage{xcolor}
\usepackage{booktabs}
\usepackage{multirow}
\usepackage{array}

\usepackage{algorithm}
\usepackage{algpseudocode}
\usepackage{makecell} 
\usepackage{booktabs,tabularx}
\usepackage{pifont}
\usepackage[table]{xcolor}

\newcommand{\rand}{{test-rand}\xspace}
\newcommand{\unseen}{{test-unseen}\xspace}
\newcommand{\shortname}{\textsc{RimRule}\xspace}

\title{R{\normalsize IM}R{\normalsize ULE}: Improving Tool-Using Language Agents\\via MDL-Guided Rule Learning}

\author{
{\bfseries
Xiang Gao\textsuperscript{1} \quad
Yuguang Yao\textsuperscript{1} \quad
Qi Zhang\textsuperscript{2} \quad
Kaiwen Dong\textsuperscript{1}
}\\
{\bfseries
Avinash Baidya\textsuperscript{1} \quad
Ruocheng Guo\textsuperscript{1} \quad
Hilaf Hasson\textsuperscript{1} \quad
Kamalika Das\textsuperscript{1}
}\\
\\
\textsuperscript{1}Intuit AI Research \\
\textsuperscript{2}Temple University \\
\\
\texttt{\{xiang\_gao, kamalika\_das\}@intuit.com}
}

\begin{document}

\maketitle

\begin{abstract}
Large language models (LLMs) often struggle to use tools reliably in domain-specific settings, where APIs may be idiosyncratic, under-documented, or tailored to private workflows. This highlights the need for effective adaptation to task-specific tools. We propose \shortname, a neuro-symbolic approach for LLM adaptation based on dynamic rule injection. Compact, interpretable rules are distilled from failure traces and injected into the prompt during inference to improve task performance. These rules are proposed by the LLM itself and consolidated using a Minimum Description Length (MDL) objective that favors generality and conciseness. Each rule is stored in both natural language and a structured symbolic form, supporting efficient retrieval at inference time. Experiments on tool-use benchmarks show that this approach improves accuracy on both seen and unseen tools without modifying LLM weights. It outperforms prompting-based adaptation methods and complements finetuning. Moreover, rules learned from one LLM can be reused to improve others, including long reasoning LLMs, highlighting the portability of symbolic knowledge across architectures.
\footnote{\emph{Proceedings of the 64th Annual Meeting of the Association for Computational Linguistics} (Volume 1: Long Papers), pages 34631–34646
July 2-7, 2026 (ACL 2026). Copyright 2026
by the authors.}

\end{abstract}

\section{Introduction}

Humans adapt by trying, erring, and compressing experience into reusable heuristics \citep{kolb1984experiential,metcalfe2017learning,gigerenzer2011heuristic}.
Such compact representation of regularities achieve cognitive economy---they preserve what matters for performance while discarding incidental detail, making knowledge easier to reuse across tasks and by other people \citep{rosch1978principles}. 


\begin{figure}[ht]
\centering
\includegraphics[width=0.85\linewidth]{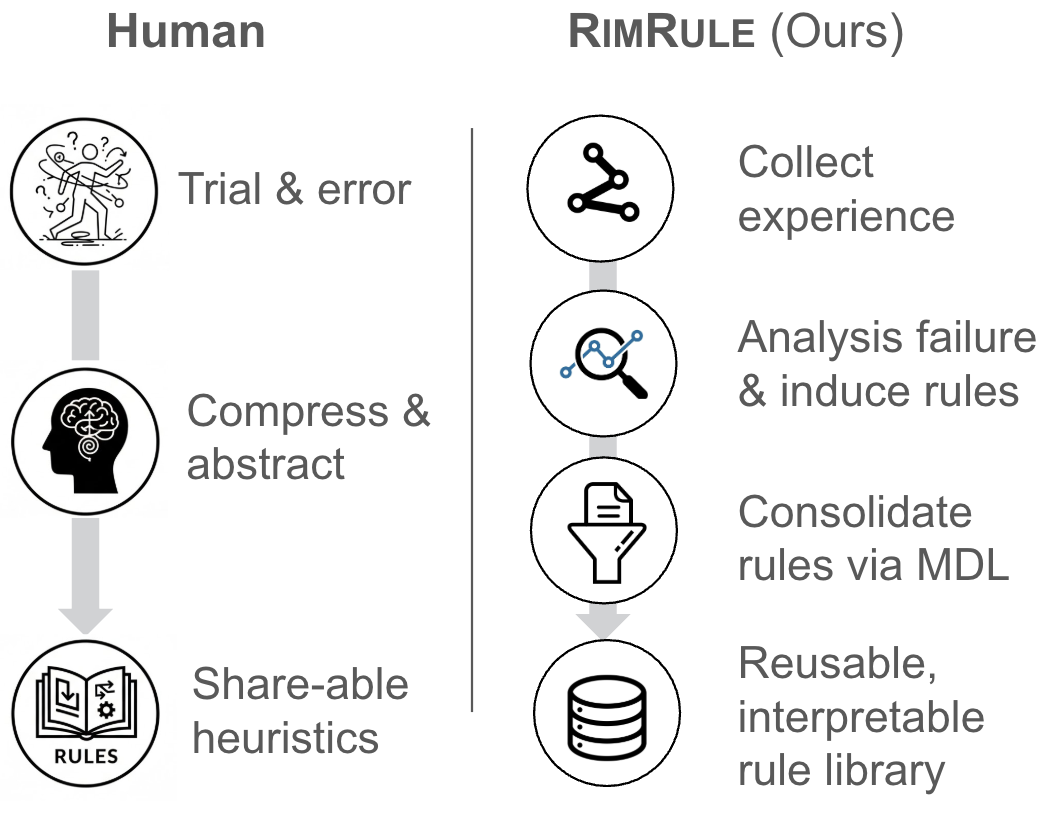}
\caption{Learning reusable and interpretable rules from experience}
\label{fig:intro}
\end{figure}

Large language models (LLMs) \citep{brown2020language, bai2023qwen, team2023gemini, grattafiori2024llama} also require adaptation, especially when deployed in unfamiliar domains with specialized tools or under-documented APIs. 
However, the way LLMs adapt differs markedly from how humans do. Today, adaptation typically takes one of three forms: (i) retrieved examples for few-shot prompting~\citep{brown2020language,yizhe2021fewshot}, (ii) globally tuned prompts~\citep{pryzant2023automatic,cui2025see}, or (iii) fine-tuned model weights~\citep{hu2021lora}.
Each approach is powerful, but none supports abstraction, reuse, and interpretability in the way humans generalize from experience. Few-shot prompting reuses raw supervision but does not abstract—and LLMs often extract only shallow patterns from examples \citep{wei2023differently}; global prompts are static and brittle in interactive settings \citep{verma2024brittle}; and tuning model weights is costly, requiring retraining whenever the environment changes, 
and preventing easy sharing of acquired knowledge across models \citep{yang2024model}.

We explore a fourth paradigm for adapting LLMs—one that mirrors how humans learn from failure, as illustrated in Fig.~\ref{fig:intro}. Instead of tuning weights or retrieving demonstrations, we induce interpretable rules. At training time, these rules are proposed in response to observed failures. 
These rules are consolidated using the Minimum Description Length (MDL) principle~\citep{rissanen1978modeling,Grunwald07MDL} and stored in a dual natural-language and symbolic representation, enabling both principled compression and reliable retrieval.
At inference time, relevant rules are dynamically retrieved and injected into the prompt to improve LLM performance. 
This constitutes a distinct adaptation paradigm: unlike few-shot prompting, it emphasizes abstraction and compression over raw example replay; unlike finetuning, it produces interpretable artifacts. Because these rules are both understandable to humans and legible to modern LLMs, they can be reused across different LLMs without retraining.


We demonstrate this approach on two tool-use benchmarks—ToolHop~\citep{ye2025toolhop} and BFCL~\citep{berkeley-function-calling-leaderboard}—where LLMs must reason over tool descriptions and invoke them correctly. Simple retry mechanisms based on tool error messages, or even the use of advanced long reasoning models, often fall short in reliably correcting systematic failures.
Our results highlight several key observations. 
\begin{itemize}
    \item First, inference-time rule injection consistently improves performance, including on queries involving tools that never appear in training.
    \item Second, learned rules transfer across LLMs in both directions: rules distilled from weaker models improve stronger, long-reasoning LLMs, and rules learned from stronger models can also benefit smaller or less capable ones.
    \item Third, our method outperforms prompting based adaptation approaches and complements finetuning, offering a distinct and synergistic axis of generalization. 
\end{itemize}


\section{Method}

\begin{figure*}[ht]
\centering
\includegraphics[width=0.95\linewidth]{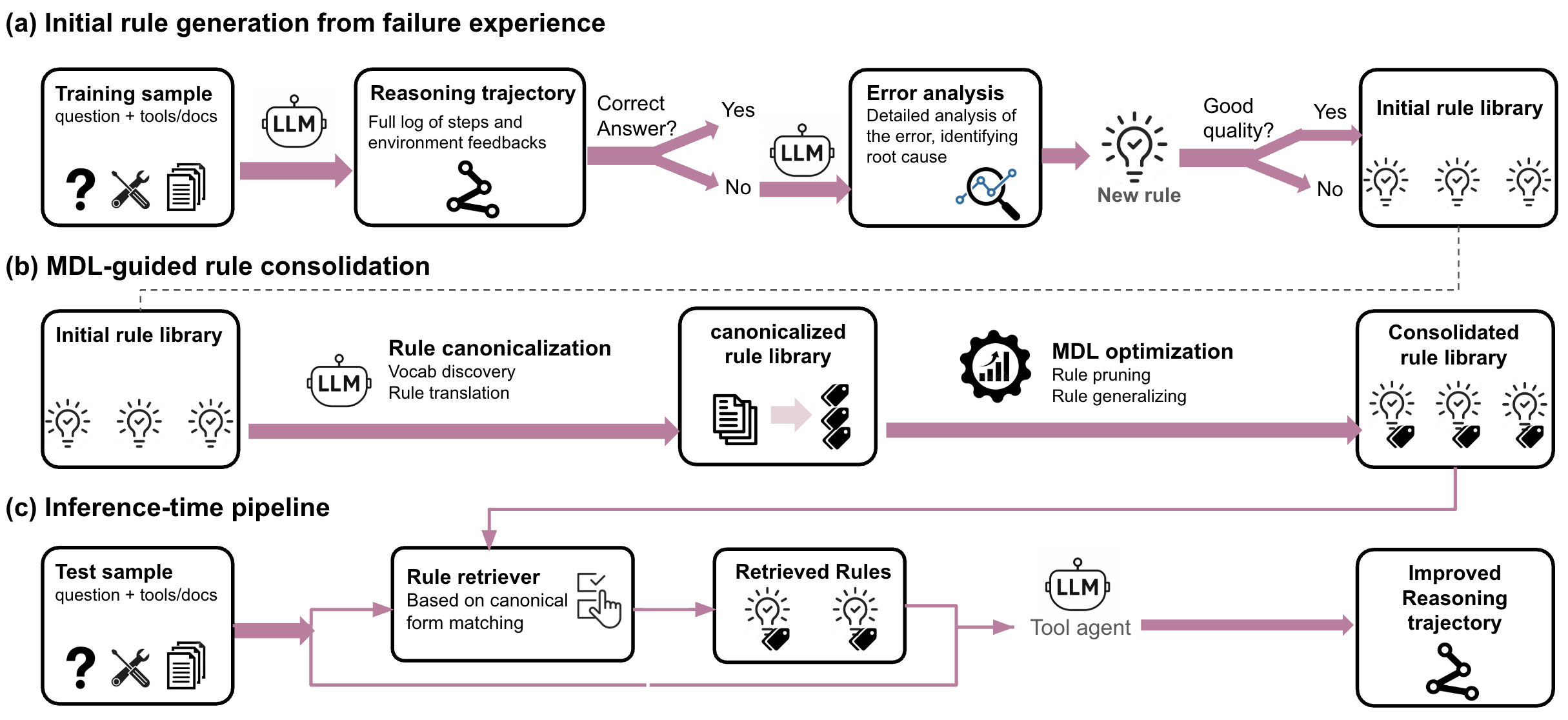}
\caption{The framework of our MDL-guided rule learning.}
\label{fig:system}
\end{figure*}

We introduce \shortname\footnote{\textbf{R}eusable, \textbf{I}nterpretable, and \textbf{M}DL-guided Rules}, a scalable framework that equips LLM-based agents with inference-time intepretable rules distilled from failure traces. 
Figure~\ref{fig:system} illustrates the overall architecture, and Table~\ref{alg:mdlrule} provides pseudocode.

Instead of learning rules sequentially—an approach that is sensitive to data order and difficult to scale—\shortname decouples rule induction into two phases. In the first stage (Section~\ref{sec:ebl}), candidate rules are generated independently from individual failures, enabling parallel, order-agnostic extraction across large interaction logs. In the second stage (Section~\ref{sec:consolidate}), these candidates are consolidated under an MDL objective, which prunes redundancy and selects a compact, high-utility rule library. This design avoids the path dependence of sequential rule learning while naturally supporting scalable rule discovery and reuse. 


\subsection{Local Rule Generation}
\label{sec:ebl}

\subsubsection{Experience Collection}

We begin by running a zero-shot LLM-based agent over each sample in the training set and
recording its execution trace. When the resulting trace produces an incorrect answer or
terminates prematurely due to tool-calling errors, we retain it as experience for rule
generation. The collection of such failure traces constitutes the experience signal used
for downstream rule induction.
Each training instance is represented as a tuple \( (x, \mathcal{S}, \tau^-, \tau^*) \),
where \( x \in \mathcal{X} \) is the user query, \( \mathcal{S} \) is the set of available
tools, \( \tau^- \) is the agent’s incorrect execution trace, and \( \tau^* \) is the
corresponding ground-truth trace.\footnote{
In settings without traces but with a reward function or score,
the same mechanism can be adapted by proposing multiple candidate rules and selecting those
that maximize observed reward on the instance; a full treatment is left to future work.}

\subsubsection{Rule Generation via EBL}

Given a failure trace \( \tau^- \) and its ground-truth counterpart \( \tau^* \), the rule
generator compares the two executions to identify a root-cause \emph{reasoning} failure,
excluding downstream propagation effects, and proposes a compact rule
\( r \in \mathcal{R}_0 \) to correct similar errors. Each rule is tagged with an error type
\( d(r) \in \{\mathrm{dec}, \mathrm{sel}, \mathrm{arg}\} \), corresponding to decomposition,
tool selection, or argument construction.

To abstract reusable rules, we follow \emph{Explanation-Based Learning} (EBL): a grounded
explanation is generated and then generalized by removing instance-specific details
\citep{mitchell1986explanation,dejong1986explanation}. The generator is implemented as an
LLM-based function.\footnote{See appendix for the prompt.}

Proposed rules are filtered by a \emph{predictive check}, requiring improved performance on
the same query when injected, and a \emph{linguistic check} enforcing a clean \emph{if--then}
form and bounded length. If a rule yields only partial improvement, the generator is
re-invoked to propose additional rules until the trace is corrected or no further atomic
fixes are found.

\begin{algorithm}[h]
\caption{\shortname}
\label{alg:mdlrule}
\small
\begin{algorithmic}[1]
\Require Training failures $D = \{(x_i, \mathcal{S}_i, \tau_i^-, \tau_i^*)\}_{i=1}^n$
\Ensure Compact symbolic rule library $\mathcal{R}$

\Statex \textbf{Stage 1: Rule Generation}
\ForAll{$(x_i, \mathcal{S}_i, \tau_i^-, \tau_i^*) \in D$ \textbf{in parallel}}
    \State $r_i^{\text{NL}} \gets \text{LLM-Propose-Rule}(\cdot)$
    \If{Passes checks}
        \State Add $r_i^{\text{NL}}$ to pool $\mathcal{R}_0$
    \EndIf
\EndFor

\Statex \textbf{Symbolic Translation}
\State Induce vocabulary $\mathcal{V}$ from $\mathcal{R}_0$
\State Translate $\mathcal{R}_0$ to symbolic rules $\mathcal{R}$

\Statex \textbf{Stage 2: MDL Consolidation}
\Repeat
    \ForAll{$r \in \mathcal{R}$}
        \State Evaluate \textsc{Prune}($r$) and \textsc{Generalize}($r$)
        \If{$\Delta L_{\text{MDL}} < 0$}
            \State Apply best edit
        \EndIf
    \EndFor
\Until{no improvement}

\State \Return $\mathcal{R}$
\end{algorithmic}
\end{algorithm}

\subsubsection{Symbolic Representation} 
\label{sec:canonical} 

Natural-language renderings of the same rule often vary in phrasing, hindering both structural matching and consistent description-length computation. We therefore compile each rule into a symbolic representation with fixed fields and closed vocabularies, enabling principled MDL consolidation (Section~\ref{sec:mdl}) and efficient retrieval (Section~\ref{sec:retrieval}). 

The symbolic schema consists of five semantic fields: \texttt{Domain} (broad topic), \texttt{Qualifier} (context), \texttt{Action} (prescribed operation), \texttt{Strength} (priority), and \texttt{ToolCategory} (abstract tool type). Each field draws values from a finite vocabulary. 

We construct these vocabularies by prompting an LLM (see Appendix~\ref{app:prompts}) over batches of natural-language rules and select the most compact vocabulary across randomized orderings. Once fixed, each rule is translated into its symbolic form by assigning valid field values, yielding a deterministic and auditable rule set.
\subsection{Rule Consolidation}
\label{sec:consolidate}

Rule generation typically produces a large set of overlapping and overly specific rules.
Consolidation is therefore required to select a compact subset that generalizes well.


We consider two approaches: a prompt-based heuristic that merges rules using an LLM, and a data-driven alternative based on the Minimum Description Length (MDL) principle.

\subsubsection{Prompt-Based}

A simple consolidation strategy is to prompt an LLM to merge or rewrite candidate rules,
as in AutoRule~\citep{wang2025autorule}. While this can reduce redundancy at the textual
level, it is heuristic and does not account for how rules affect agent behavior on actual
failure cases. This motivates a consolidation objective that explicitly trades off rule
complexity against empirical effectiveness.

\subsubsection{MDL-Guided}
\label{sec:mdl}

We adopt the \emph{Minimum
Description Length} (MDL) principle, which favors models that compress observed data while
remaining concise~\citep{rissanen1978modeling,Grunwald07MDL}. In our setting, MDL provides
a principled objective for selecting a compact rule set that corrects as many failures as
possible without unnecessary complexity.

\paragraph{Objective.}
Let \( H \subseteq \mathcal{R}^{\text{sym}} \) denote a candidate rule library. We minimize
\[
\mathrm{MDL}(H) = L(H) + L(D \mid H),
\]
where \( L(H) \) penalizes rule complexity and \( L(D \mid H) \) measures how well the rules
correct observed failures.

\paragraph{Model Cost.}
We define a length-based prior over rule libraries,
\[
P(H) \propto \exp\!\Bigl(-\alpha \sum_{r \in H} \ell(r)\Bigr),
\]
where \( \ell(r) \) denotes the \textbf{symbolic token length}
\footnote{Symbolic token length provides a canonical measure of rule complexity by
abstracting away incidental natural-language phrasing.}
of rule \( r \) and \( \alpha > 0 \)
controls the strength of regularization. This corresponds to a maximum-entropy prior under
a constraint on expected total rule length~\citep{jaynes1957maxent}. Taking the negative
log-likelihood yields the model code length
\[
L(H) = -\log P(H) = \alpha \sum_{r \in H} \ell(r) + \log Z(\alpha),
\]
where \( Z(\alpha) \) is constant for a fixed candidate pool and is therefore omitted during
optimization.

\paragraph{Data Cost.}
Given failure cases \( D = \{(x_i, \mathcal{S}_i, \tau_i^-, \tau_i^*)\}_{i=1}^n \), we define
\( a_i(H) = 1 \) if injecting \( H \) corrects failure \( i \). Let
\( k_H = \sum_i a_i(H) \). We model correction outcomes with a Bernoulli likelihood and
use the refined MDL plug-in code
\[
L(D \mid H) = -\bigl[k_H \log \hat{p}_H + (n-k_H)\log(1-\hat{p}_H)\bigr],
\]
where \( \hat{p}_H = k_H/n \). This term directly rewards rule sets that correct more failures.

\paragraph{Greedy Consolidation.}
We minimize \( \mathrm{MDL}(H) \) starting from the full rule pool and applying local edits
that strictly reduce the objective. We consider two operations: \emph{pruning} a rule, and
\emph{generalizing} a tool-specific rule to a category-level rule. Edits are accepted only
when the reduction in model cost outweighs any increase in data cost. This greedy procedure
terminates when no local edit improves the MDL objective.

In practice, consolidation scales linearly with the size of the candidate rule pool, as each iteration evaluates a constant set of local edits per rule. While this greedy procedure does not guarantee a globally optimal solution and may miss beneficial non-local edits, it follows standard practice in rule learning and MDL-based pruning, where local edits are commonly used to remove redundancy and over-specific rules and are empirically effective in improving generalization~\citep{Quinlan87Simplifying,Furnkranz97Pruning,Grunwald07MDL}. Exploring more global or approximate inference strategies for rule consolidation is an interesting direction for future work.

\paragraph{MDL Behavior.}
Although the term \( L(D \mid H) \) could be minimized by correcting no failures,
this degenerate case is never reached in practice. Consolidation proceeds greedily from a
high-performing initial rule set; in this regime, removing a rule that reduces the number
of corrected failures sharply increases \( L(D \mid H) \), outweighing the modest reduction
in model cost. Such edits are therefore rejected.

\paragraph{Choice of $\alpha$.}
We select the regularization strength \( \alpha \) by choosing the value that yields the best
end-to-end accuracy after consolidation on the failure set. Since each rule is evaluated on
failures beyond its originating instance, this selection already reflects cross-sample
generalization. Performance is stable across a wide range of \( \alpha \), indicating that
MDL acts as a robust regularizer.

\subsection{Rule Retrieval}
\label{sec:retrieval}

At inference time, injecting all learned rules is inefficient and noisy, making retrieval
necessary to select only rules relevant to the current query and tool context.

\subsubsection{Natural language based}

A simple baseline retrieves rules directly from their natural-language form by encoding the
query and available tools with an off-the-shelf sentence embedding model and selecting the
top-$k$ rules by cosine similarity. While easy to implement, this approach relies on generic
embeddings that struggle with lexical variability and fine-grained rule semantics. Finetuning
a dedicated embedding model introduces additional complexity, including limited training
data, the need for ground-truth rule annotations, and reduced scalability.

\subsubsection{Symbolic-based}

To address these limitations, and inspired by prior work on structure-aware retrieval
\citep{munikoti2023atlantic,li2023santa}, we perform retrieval over the symbolic
representation of rules. By structuring both rules and queries into shared semantic fields,
symbolic matching is more efficient, interpretable, and more robust to incidental linguistic
variation than retrieval over raw text.

Given a query \( x_q \) and available tools \( \mathcal{S}_q \), we first convert the state into a symbolic representation \( z_q \). Retrieval proceeds in two stages. First, a coarse filter removes inapplicable rules: decomposition rules are always retained, while tool-use rules are filtered based on whether their scope matches the available tools or tool categories. Second, the remaining rules are ranked by semantic similarity between their symbolic fields and those of \( z_q \), using a generic embedding model. We retain the top-$k$ rules and inject them into the prompt in natural-language form. Unlike retrieval over raw text, similarity is computed over a fixed symbolic vocabulary, where NL-to-symbolic conversion has already resolved much of the lexical and structural variation that generic embeddings struggle to capture.

\begin{table}[h]
\small
\caption{Number of examples in each dataset split.}
\vspace{0.5em}
\label{tab:dataset}
\centering
\renewcommand{\arraystretch}{1.25}
\begin{tabular}{@{}l|c|cc@{}}
\toprule
\multirow{2}{*}{\textbf{Dataset}} & 
\multirow{2}{*}{\textbf{Train}} & 
\multicolumn{2}{c}{\textbf{Test}} \\
 &  & \textbf{Rand} & \textbf{Unseen} \\
\midrule
ToolHop & 392 & 70 & 51 \\
BFCL:Live-Multiple & 735 & 175 & 143 \\
BFCL:Multi-Turn-Base & 90 & 60 & 50 \\
\bottomrule
\end{tabular}
\end{table}

\begin{table*}[t]
\small
\caption{{Accuracy ($\pm$ standard deviation, in \%) and number of rules at each learning stage.} Rule consolidation improves performance while reducing rule count.}
\vspace{0.5em}
\label{tab:learning}
\centering
\renewcommand{\arraystretch}{1.25}
\begin{tabular}{@{}l|ccc|ccc@{}}
\toprule
 & \multicolumn{3}{c|}{\textbf{ToolHop}} & \multicolumn{3}{c}{\textbf{BFCL}} \\
 & Test-rand & Test-unseen & \# Rules & Test-rand & Test-unseen & \# Rules \\
\midrule
Initial
& $26.5 \scriptstyle \pm 1.3$ & $35.1 \scriptstyle \pm 1.6$ & 0
& $50.1 \scriptstyle \pm 0.9$ & $45.0 \scriptstyle \pm 1.0$ & 0 \\
+ Rule Gen.
& $27.6 \scriptstyle \pm 1.3$ & $42.1 \scriptstyle \pm 1.6$ & 72
& $54.8 \scriptstyle \pm 1.3$ & $47.6 \scriptstyle \pm 1.4$ & 151 \\
+ Consolid.
& $\mathbf{31.1} \scriptstyle \pm 1.3$ & $\mathbf{43.1} \scriptstyle \pm 1.6$ & 67
& $\mathbf{56.6} \scriptstyle \pm 1.2$ & $\mathbf{48.5} \scriptstyle \pm 1.4$ & 121
\\
\bottomrule
\end{tabular}
\end{table*}

\begin{table*}[h]
\small
\caption{{Accuracy ($\pm$ standard deviation) of LLMs on Toolhop and BFCL, with and without rules.} Rules are learned once from Llama3.2 and reused across models.}
\vspace{0.5em}
\label{tab:llm_rule_reuse}
\centering
\renewcommand{\arraystretch}{1.25}
\begin{tabular}{@{}>{\centering\arraybackslash}m{1.8cm}|>{\centering\arraybackslash}m{1.7cm}|cc|cc@{}}
\toprule
\multirow{2}{=}{\makecell[c]{\textbf{Learned}\\ \textbf{from}}} & \multirow{2}{=}{\makecell[c]{\textbf{Applied}\\ \textbf{on}}} & \multicolumn{2}{c|}{\textbf{ToolHop}} & \multicolumn{2}{c}{\textbf{BFCL}} \\
& & No Rules & \shortname & No Rules & \shortname \\
\midrule
\multirow{4}{*}{\centering Llama3.2}
& Llama3.2 & $26.5 \scriptstyle \pm 1.3$ & $\mathbf{31.1} \scriptstyle \pm 1.3$
& $50.1 \scriptstyle \pm 0.9$ & $\mathbf{56.6} \scriptstyle \pm 1.2$ \\
& GPT-4o & $58.1 \scriptstyle \pm 1.4$ & $57.4 \scriptstyle \pm 1.4$
& $71.6 \scriptstyle \pm 0.8$ & $\mathbf{75.6} \scriptstyle \pm 1.1$ \\
& Llama4 & $73.8 \scriptstyle \pm 1.2$ & $\mathbf{76.7} \scriptstyle \pm 1.2$
& $75.8 \scriptstyle \pm 0.8$ & $\mathbf{77.5} \scriptstyle \pm 1.1$ \\
& O1 & $53.2 \scriptstyle \pm 1.4$ & $\mathbf{57.4} \scriptstyle \pm 1.4$
& $75.6 \scriptstyle \pm 0.8$ & $\mathbf{77.6} \scriptstyle \pm 1.1$ \\

\midrule

\multirow{4}{*}{\centering GPT-4o}
& Llama3.2 & $26.5 \scriptstyle \pm 1.3$ & $\mathbf{31.3} \scriptstyle \pm 1.3$
& $50.1 \scriptstyle \pm 0.9$ & $\mathbf{52.4} \scriptstyle \pm 1.3$ \\
& GPT-4o & $58.1 \scriptstyle \pm 1.4$ & $\mathbf{60.3} \scriptstyle \pm 1.4$
& $71.6 \scriptstyle \pm 0.8$ & $\mathbf{76.4} \scriptstyle \pm 1.1$ \\
& Llama4 & $73.8 \scriptstyle \pm 1.2$ & $\mathbf{77.4} \scriptstyle \pm 1.2$
& $75.8 \scriptstyle \pm 0.8$ & $\mathbf{77.7} \scriptstyle \pm 1.0$ \\
& O1 & $53.2 \scriptstyle \pm 1.4$ & $\mathbf{56.1} \scriptstyle \pm 1.2$
& $75.6 \scriptstyle \pm 0.8$ & $\mathbf{77.9} \scriptstyle \pm 1.1$ \\

\bottomrule
\end{tabular}
\end{table*}

\begin{table*}[t]
\small
\caption{Comparison of prompting-based adaptation methods on ToolHop and BFCL.}

\vspace{0.4em}
\label{tab:vs_prompt}
\centering
\setlength{\tabcolsep}{8pt}
\renewcommand{\arraystretch}{1.12}
\begin{tabular}{@{}c|cc|cc@{}}

\toprule 
& \multicolumn{2}{c|}{\textbf{ToolHop}} & \multicolumn{2}{c}{\textbf{BFCL}} 
\\ 
& Test-rand & Test-unseen & Test-rand & Test-unseen \\ 
\midrule

Zero-shot            
& $26.5 \scriptstyle ~\pm 1.3$ & $35.1 \scriptstyle \pm 1.6$  
& $50.1 \scriptstyle ~\pm 0.9$ & $45.0 \scriptstyle \pm 1.0$ 
 \\

Few-shot \citep{yizhe2021fewshot}           
& $29.9 \scriptstyle ~\pm 1.4$ & $37.9 \scriptstyle ~\pm 1.4$  
& $54.5 \scriptstyle ~\pm 0.9$ & $46.6 \scriptstyle ~\pm 1.4$ 
 \\

SEE \citep{cui2025see} 
& $27.6 \scriptstyle ~\pm 1.5$ & $35.9 \scriptstyle ~\pm 1.5$    
& $52.2 \scriptstyle ~\pm 1.0$ & $45.5 \scriptstyle ~\pm 1.0$ 
 \\

\shortname (Ours)  
& $\mathbf{31.1} \scriptstyle \pm 1.3$ & $\mathbf{43.1} \scriptstyle \pm 1.6$ 
& $\mathbf{56.6} \scriptstyle \pm 1.2$ & $\mathbf{48.5} \scriptstyle \pm 1.4$
 \\

\bottomrule
\end{tabular}
\end{table*}

\begin{table*}[t]
\small

\caption{Performance with and without \shortname for finetuned models.}

\vspace{0.4em}  
\label{tab:vs_finetune}
\centering
\setlength{\tabcolsep}{8pt}
\renewcommand{\arraystretch}{1.12}
\begin{tabular}{@{}l|cc|cc@{}}
\toprule
             & \multicolumn{2}{c|}{\textbf{ToolHop}} & \multicolumn{2}{c}{\textbf{BFCL}} \\ 
\cmidrule(lr){2-3} \cmidrule(lr){4-5}
             & Test-rand & Test-unseen & Test-rand & Test-unseen \\ 
             
\midrule

SFT (\texttt{Llama3.2})
& $43.8 \scriptstyle ~\pm 1.4$ & $38.5 \scriptstyle ~\pm 1.5$
& $65.0 \scriptstyle ~\pm 0.9$ & $58.7 \scriptstyle ~\pm 1.4$ \\ 

+ \shortname          
& $\mathbf{50.0} \scriptstyle ~\pm 1.4$ & $\mathbf{45.1} \scriptstyle ~\pm 1.6$
& $\mathbf{68.6} \scriptstyle ~\pm 1.2$ & 
$\mathbf{62.7} \scriptstyle ~\pm 1.3$    
\\ 

\midrule

Function Calling  (\texttt{GPT4o})        
& $63.8\scriptstyle ~\pm 1.4$ & $ 83.0 \scriptstyle ~\pm 1.2$
& $79.6 \scriptstyle ~\pm 0.7$ & $77.2 \scriptstyle ~\pm 0.8$ \\ 

+ \shortname          
& $\mathbf{66.1} \scriptstyle ~\pm 1.4$ & $\mathbf{85.4} \scriptstyle ~\pm 1.1$
& $\mathbf{81.7} \scriptstyle ~\pm 1.0$ & $\mathbf{79.8} \scriptstyle ~\pm 1.0$ 
\\

\bottomrule
\end{tabular}
\end{table*}

\section{Experiments}

\subsection{Datasets}

We evaluate our method on two tool-use benchmarks: \textbf{ToolHop}\footnote{\url{https://hf.co/datasets/bytedance-research/ToolHop}, CC BY 4.0 license}~\cite{ye2025toolhop} and
\textbf{BFCL}\footnote{\url{https://hf.co/datasets/gorilla-llm/Berkeley-Function-Calling-Leaderboard}, Apache license 2.0}~\cite{berkeley-function-calling-leaderboard}. ToolHop consists of compositional,
multi-turn queries requiring multi-step tool reasoning, while the \texttt{live-multiple}
subset of BFCL provides a complementary single-step setting with a larger and more diverse
tool set, making tool selection more challenging.

For both datasets, we assume access to ground-truth execution traces during training, but no
ground-truth rules are provided. To evaluate generalization, we define two splits:
\textbf{\rand}, a random in-distribution split, and \textbf{\unseen}, a held-out split
containing queries with tools not seen during training. Dataset statistics are summarized
in Table~\ref{tab:dataset}.

\subsection{Baselines}

We compare \shortname against several established adaptation paradigms:
\begin{itemize}
    \item \textbf{Few-shot in-context learning} retrieves top-$k$ training examples based on
    semantic similarity~\citep{yizhe2021fewshot}.
    \item \textbf{Prompt optimization} with SEE~\citep{cui2025see}, which performs global prompt
    optimization by jointly evolving instructions and demonstrations.
    \item \textbf{Finetuning}, including supervised finetuning (SFT) of open-source models using
    LoRA~\citep{hu2021lora}, as well as closed-source models tuned for function calling.
    For SFT, we construct training data from ground-truth execution traces, training the model
    to generate correct tool selections and arguments conditioned on the query, available tools,
    and previous tool-call results.
\end{itemize}


All methods, including ours, use a \textbf{ReAct}-style prompting framework~\citep{yao2023react} for tool-use
reasoning, allowing agents to generate intermediate steps and tool calls interactively.
We permit agents to retry based on tool feedback or error messages, a common robustness
mechanism that nonetheless struggles with systematic reasoning errors. Our method builds
on this setup by injecting explicit rules to prevent such failures.

We use short names for evaluated models: \texttt{Llama3.2}\footnote{\texttt{meta.llama3-2-3b-instruct-v1-0}},
\texttt{Llama4}\footnote{\texttt{meta.llama4-maverick-17b-instruct-v1-0}},
\texttt{GPT-4o}\footnote{\texttt{gpt-4o-2024-11-20}}, and \texttt{O1}\footnote{\texttt{o1-mini-2024-09-12}},
where \texttt{O1} is a long-reasoning model.

The symbolic representation supports rule consolidation and retrieval, not
direct symbolic execution. Because the tool-use tasks we study involve noisy natural-language
queries and environments, purely symbolic rule application is brittle; accordingly, we do not
compare against symbolic rule learners designed for clean, structured data\citep{quinlan1990foil, clark1989cn2, Cohen95, Furnkranz97Pruning}. Instead, rules are applied in natural-language form
to guide LLM reasoning.



\section{Results and Discussion}

\subsection{Learning Process}

Table~\ref{tab:learning} illustrates the learning trajectory using Llama3.2 as a representative
example. Other models exhibit similar trends. 

Starting from an empty rule
library, we collect failure traces from a zero-shot agent and generate candidate rules, yielding 72 rules on ToolHop and 151 on BFCL\footnote{See Appendix for sample rules}. 
This initial rule set improves accuracy on both \rand and
\unseen splits. 

Applying MDL-guided consolidation further reduces the rule count (by 7\% on ToolHop and 20\% on BFCL) while
improving performance on \rand and yielding smaller but consistent gains on \unseen.
Overall, the process produces a compact, interpretable rule library that improves both
in-distribution and out-of-distribution generalization.

\subsection{Case Study}

Table~\ref{tab:case-study} shows a representative failure where the agent directly queries a
complex familial relationship, leading to a tool error. \shortname identifies the root cause
as improper decomposition and induces a rule that enforces stepwise resolution of intermediate
entities. The resulting rule is abstract and reusable, and its symbolic form enables reliable
retrieval for similar queries. This example illustrates how failure traces are converted into
compact rules that correct systematic reasoning errors beyond the original instance.

\begin{table}[t]
\centering
\small
\caption{Case study illustrating failure-driven rule induction and symbolic representation.}
\label{tab:case-study}
\begin{tabular}{p{0.18\linewidth} p{0.76\linewidth}}
\toprule
\textbf{Query} &
How many letters (excluding the first and last) are there in the first name of
Viacheslav~I of Kiev's maternal grandfather? \\
\midrule
\textbf{Failure} &
The agent directly queries \textit{Viacheslav~I of Kiev's maternal grandfather}.
The tool returns an error: \textit{``no data found''}. \\
\midrule
\textbf{Proposed Rule} &
If the user query involves identifying a specific familial relationship (e.g., maternal
grandfather), then decompose the task by first resolving intermediate relationships
(e.g., mother, father) sequentially. \\
\midrule
\textbf{Symbolic Form} &
\texttt{if} \;
\texttt{domain=FAMILIAL\_RELATIONSHIP} \;\textbf{or}\;
\texttt{tool\_category=GENEALOGY\_QUERY}, \\
& \texttt{then action=[DECOMPOSE\_QUERY, RESOLVE\_INTERMEDIATE\_ENTITY, SEQUENCE\_SUBTASKS]} \\
& \texttt{strength=MANDATORY} \\
\bottomrule
\end{tabular}
\end{table}

\subsection{Re-usability Across LLMs}

Humans can share heuristics because they are interpretable across contexts. By encoding
adaptation as symbolic, human-readable rules, our method enables knowledge to be reused
across LLMs without retraining.

We learn two rule libraries from failure traces produced by \texttt{Llama3.2} and
\texttt{GPT-4o}, and apply them to models of varying size and reasoning strength, evaluating
performance on the \rand split. Table~\ref{tab:llm_rule_reuse} shows consistent gains in both
directions: \textbf{strong-to-weak} transfer improves smaller models, while
\textbf{weak-to-strong} transfer benefits strong reasoning models such as \texttt{O1} and
\texttt{Llama4}. These results suggest that symbolic rules capture transferable failure
patterns not resolved by scale alone.

\begin{table}[t]
\small
\caption{Test-rand accuracy ($\pm$ standard deviation, in \%) comparing retrieval strategies.}
\vspace{0.5em}
\label{tab:retriever}
\centering
\renewcommand{\arraystretch}{1.25}
\begin{tabular}{@{}lcc@{}}
\toprule
\textbf{Method} & \textbf{ToolHop} & \textbf{BFCL} \\
\midrule
Nat.\ Lang.\ based 
& $29.4 \scriptstyle \pm 1.3$ 
& $54.1 \scriptstyle \pm 1.0$ \\
Symbolic guided 
& $\mathbf{31.1} \scriptstyle \pm 1.3$ 
& $\mathbf{56.6} \scriptstyle \pm 1.2$ \\
\bottomrule
\end{tabular}
\end{table}

\subsection{Performance on Small Dataset}

We evaluate \shortname in a low-resource setting on the \texttt{multi\_turn\_base} split of
BFCL using only 90 training samples. Despite limited supervision, the method learns just
four rules (Appendix~\ref{app:sample_rules}) yet achieves substantial gains: accuracy
improves from $55.2\%$ to $62.1\%$ on \rand and from $46.0\%$ to $60.0\%$ on \unseen. These
results highlight the sample efficiency and practical applicability of rule-based
adaptation.

\subsection{Outperforming Prompting-based Methods}

We compare \shortname against prompting-based adaptation methods, including ReAct
prompting \citep{yao2023react}, few-shot in-context learning \citep{yizhe2021fewshot}, and prompt optimization with SEE \citep{cui2025see}.
As shown in Table~\ref{tab:vs_prompt}, \shortname consistently outperforms all prompting
baselines on both ToolHop and BFCL, across \rand and \unseen splits, demonstrating stronger
generalization than fixed-prompt approaches.

\subsection{Complementing Finetuned Models}

As shown in Table~\ref{tab:vs_finetune}, adding \shortname to finetuned models yields consistent
gains, especially on the \unseen split, indicating that symbolic rules correct residual
systematic errors left by supervised training. We observe similar improvements when applying
\shortname to models with native function-calling capabilities (Table~\ref{tab:vs_finetune}),
showing that rule-based adaptation operates along an orthogonal axis of generalization.

\begin{table}[t]
\small
\caption{Test-rand accuracy ($\pm$ standard deviation, in \%) comparing consolidation strategies.}
\vspace{0.5em}
\label{tab:consolid}
\centering
\renewcommand{\arraystretch}{1.25}
\begin{tabular}{@{}lcc@{}}
\toprule
\textbf{Method} & \textbf{ToolHop} & \textbf{BFCL} \\
\midrule
Prompt-based 
& $27.5 \scriptstyle \pm 1.4$
& $52.1 \scriptstyle \pm 0.9$ \\
MDL-guided 
& $\mathbf{31.1} \scriptstyle \pm 1.3$ 
& $\mathbf{56.6} \scriptstyle \pm 1.2$ \\
\bottomrule
\end{tabular}
\end{table}

\subsection{Ablation Studies}

\paragraph{Impact of retrieval method.}
Table~\ref{tab:retriever} shows that symbolic retrieval consistently outperforms retrieval
based on raw natural-language similarity on both datasets.

\paragraph{Impact of consolidation method.}

As shown in Table~\ref{tab:consolid}, MDL-guided consolidation significantly outperforms
prompt-based rule merging. Together, these results indicate that both symbolic retrieval
and MDL-based consolidation are essential to our approach.

\section{Related Work}

\paragraph{Rule Learning.}
Recent work studies rule learning with LLMs via prompting, extracting rules from
demonstrations~\cite{gao2024customizing}, graphs~\cite{chen2024rulerag}, or traces~\cite{zhang2024ruag}
to guide generation~\cite{zhou2024self,wang2024ulogic}. Other approaches use rules at training
time for synthetic supervision~\cite{morishita2024enhancing}, reward modeling~\cite{wang2025autorule},
or distillation~\cite{sadeq2025improving}, or store rules in memory~\cite{wang2024symbolic}. In
contrast to these largely static or training-time uses, we learn and consolidate rules from
failures for dynamic inference-time reuse.

\paragraph{Tool-use Agents.}
Tool-use agents often rely on retriever--generator pipelines or finetuning to select and call
tools~\citep{qin2023toolllm,patil2023gorilla}, or train models to use tools via self-supervision
or agent tuning~\citep{schick2023toolformer}. Recent work further explores reinforcement
learning and agent-centric optimization for tool use~\citep{le2025toolbrain,zhang2025toolr1},
as well as generative approaches that tokenize tools for unified retrieval and calling
\citep{hao2023toolkengpt,wang2024toolgen}. While effective, these approaches primarily encode
adaptation in model parameters. We instead externalize adaptation as a compact, interpretable
rule library, enabling modular reuse across LLMs without retraining.

\section{Conclusion}

We propose \shortname, an approach that adapts LLMs by distilling compact, interpretable rules
from failure traces and applying them at inference time. Without modifying model weights, our
method improves both in- and out-of-distribution performance, outperforms prompting-based
methods, and complements finetuned models. These results position inference-time rule learning
as a practical path toward modular and interpretable LLM adaptation.

\section{Limitations}

Our approach assumes access to failure signals during training, such as execution traces or
reliable performance feedback, which may not be available in all deployment settings. Rule
consolidation is performed using a greedy MDL-based procedure; while efficient and effective
at the scales studied, it does not guarantee a globally optimal rule set, and alternative
optimization methods deserve future study.
In addition, the symbolic representation relies on a manually designed
schema intended for noisy natural-language tool-use tasks; other domains may benefit from
different symbolic abstractions. 
Finally, the effectiveness
of learned rules depends on the quality and diversity of observed failures, and gains may be
smaller in domains with sparse or highly idiosyncratic errors.

\paragraph{Potential Risks.}
As with any adaptive system, learned rules could overgeneralize or reflect biases present in
observed failures if applied outside their intended scope. We mitigate such risks by keeping
rules interpretable and applied at inference time, enabling auditing, revision, or removal.
Future work may further explore safeguards for monitoring and validating rule usage in
high-stakes settings.

\bibliography{custom}
\clearpage
\appendix
\section{Appendix}

\subsection{Prompts}
\label{app:prompts}

This section includes the prompts used throughout our pipeline: for generating rules from failure traces, discovering and updating the classification vocabulary, and translating rules into canonical form. Prompts are shown as used, with placeholders for query, tools, and traces.

\lstdefinestyle{promptstyle}{
  basicstyle=\ttfamily\small,
  breaklines=true,
  breakatwhitespace=true,
  columns=fullflexible,
  frame=single,
  backgroundcolor=\color{gray!5},
rulecolor=\color{red!50!black},
  captionpos=b
}
\lstset{inputencoding=utf8}

\begin{lstlisting}[style=promptstyle, caption={Prompt used to generate reasoning rules from failure traces. Placeholders like user query and tool schemas are injected at runtime.}]
You are a senior AI systems instructor tasked with helping a tool-using language model, referred to as the Tool Agent, improve its ability to use tools accurately, efficiently, and reliably, even with new tools and unseen queries that are structurally and/or semantically similar to previously seen examples. Your role is to review incorrect tool usage traces produced by the Tool Agent, compare them against the correct (groundtruth) traces, and analyze any embedded error messages to reason about the error.

An incorrect trace occurs when a query's execution by the Tool Agent doesn't match the groundtruth answer. Incorrect traces are either complete (finished but incorrect) or incomplete due to unrecoverable execution errors. You will be provided with the Tool Agent's full execution trace, including embedded error messages.

There are two types of errors:
i. Reasoning errors (to analyze and generate rules for)
ii. Propagation errors (to ignore, as they result from prior mistakes)

Steps may contain: 
(i) no errors, (ii) only propagation errors, or (iii) one or more reasoning errors.

Your goal is to extract generalizable, tool-schema-aware, and context-sensitive rules that can guide future tool usage-even with unseen queries and schemas. Think like a meta-cognitive teacher: diagnose root causes and articulate precise, reusable abstractions.

You will now be given:
1. The user query              [INSERT USER QUERY]
2. The available tools         [INSERT TOOL SCHEMA JSON]
3. The incorrect trace         [INSERT TOOL AGENT TRACE]
4. The groundtruth trace       [INSERT GROUNDTRUTH TRACE]

Your task is to identify reasoning errors by comparing the traces.
Check:
- Were subtasks identified correctly?
- Were the correct tools selected?
- Were the arguments constructed correctly?

Guidelines for rule generation:
- Only generate rules for reasoning errors
- Identify root causes, not superficial fixes
- Rules must be generalizable and schema-aware
- Rules must not contain query-specific or tool-specific tokens
- Each rule must be atomic and classifiable as:
  a. decomposition error
  b. tool selection error
  c. tool arguments error
- If multiple reasoning failures occur, generate multiple rules unless clarity is preserved

Rules must be symbolic, composable, and conform to a standard form (e.g., condition => action).

Output format:
First, write a brief analysis explaining the root cause and error type.
Then output one rule only, using this JSON structure:
{
  "new_rule": "<generalized if-then rule>",
  "error_type": "<decomposition error / tool selection error / tool arguments error>"
}
\end{lstlisting}

\begin{lstlisting}[style=promptstyle, caption={Prompt used to induce the rule classification vocabulary from a batch of rules. Placeholders are filled dynamically.}]

You are an expert rule classification specialist.
Your task is to CREATE the initial vocabulary for rule classification based on the provided field definitions.
Analyze the rules carefully and create comprehensive categories that cover all rule types.
Use only uppercase letters, numbers, and underscores for token names.
Return only valid JSON format.

Create initial vocabulary for rule classification based on these definitions:

FIELD DEFINITIONS:

DOMAIN (enum): Broad topical domain, inferred jointly from rule content + tool context + query context.

QUALIFIER (enum): Fine-grained situational tags that are more specific than domain.

ACTION (enum): The action(s) the rule prescribes, from a closed set.

STRENGTH (enum): Rule action priority (must, may). Same scope/domain/qualifier, higher priority should be retrieved rather than low priority.
Examples: MANDATORY, RECOMMENDED, OPTIONAL

TOOL_CATEGORY (enum): Tool functional category based on tool capabilities and usage patterns.
This should represent broad tool types that can group similar tools together.
Examples: DATA_PROCESSING, SEARCH_ENGINE, COMPUTATION, TEXT_PROCESSING...

Rules to analyze:
[INSERT_BULLETS_HERE]

Create a comprehensive vocabulary covering all rule types in the above rules.
Use only uppercase letters, numbers, and underscores for category names.

Return JSON:
{
  "vocab_version": "v1",
  "domain": [ ... ],
  "qualifier": [ ... ],
  "action": [ ... ],
  "strength": [ ... ],
  "tool_category": [ ... ]
}
\end{lstlisting}

\begin{lstlisting}[style=promptstyle, caption={Prompt used to update and expand the rule classification vocabulary using newly generated rules.}]
You are an expert rule classification specialist.
Your task is to UPDATE and EXPAND an existing vocabulary based on new rules.
Review the current vocabulary, identify missing categories from new rules, and return the complete updated vocabulary.
Use only uppercase letters, numbers, and underscores for token names.
Avoid duplicates and maintain consistency with existing categories.
Return only valid JSON format.

Update and expand vocabulary for rule classification.

FIELD DEFINITIONS:

DOMAIN (enum): Broad topical domain, inferred jointly from rule content + tool context + query context.

QUALIFIER (enum): Fine-grained situational tags that are more specific than domain.

ACTION (enum): The action(s) the rule prescribes, from a closed set.

STRENGTH (enum): Rule action priority (must, may). Same scope/domain/qualifier, higher priority should be retrieved rather than low priority.

TOOL_CATEGORY (enum): Tool functional category based on tool capabilities and usage patterns.

Current accumulated vocabulary:
DOMAIN: [INSERT CURRENT DOMAINS OR 'None yet']
QUALIFIER: [INSERT CURRENT QUALIFIERS OR 'None yet']
ACTION: [INSERT CURRENT ACTIONS OR 'None yet']
STRENGTH: [INSERT CURRENT STRENGTHS OR 'None yet']
TOOL_CATEGORY: [INSERT CURRENT TOOL_CATEGORIES OR 'None yet']

New rules to analyze:
[INSERT NEW RULE BULLETS HERE]

Instructions:
1. Review the current vocabulary above
2. Analyze the new rules to identify any missing categories
3. Add new categories if needed, but avoid duplicates
4. Keep existing categories that are still relevant
5. Return the complete updated vocabulary

Return JSON with all categories (existing + new):
{
  "domain": ["existing_domains", "new_domains_if_any"],
  "qualifier": ["existing_qualifiers", "new_qualifiers_if_any"],
  "action": ["existing_actions", "new_actions_if_any"],
  "strength": ["existing_strengths", "new_strengths_if_any"],
  "tool_category": ["existing_tool_categories", "new_tool_categories_if_any"]
}
\end{lstlisting}

\begin{lstlisting}[style=promptstyle, caption={Prompt used to classify individual rules using a fixed vocabulary. Vocabulary values and rule content are injected at runtime.}]
You are an expert rule classification specialist.
Your task is to classify individual rules using the provided vocabulary.
Analyze the rule content, tool context, and query context to determine the most appropriate classification.
Choose only from the given vocabulary options.
Return only valid JSON format.

Vocab enumerations:
- domain: [INSERT domain list]
- qualifier: [INSERT qualifier list]
- action: [INSERT action list]
- strength: [INSERT strength list]
- tool_category: [INSERT tool_category list]

Rule to classify (analyze all fields for context):
[INSERT RULE OBJECT AS JSON]

Classification Guidelines:

DOMAIN: Infer from rule content + tool context + query context. Choose the broadest applicable domain that covers the rule's topic.

QUALIFIER: Select fine-grained situational tags that are more specific than domain. You can select multiple qualifiers if the rule applies to multiple situations. Focus on conditions, constraints, or specific contexts mentioned in the rule.

ACTION: Identify the specific action(s) the rule prescribes. Choose from the closed set of available actions. You can select multiple actions if the rule prescribes multiple steps.

STRENGTH: Determine rule priority (must, may). MANDATORY = must follow, RECOMMENDED = should follow, OPTIONAL = may follow. Consider the rule's language and context to determine priority.

TOOL_CATEGORY: Identify the functional category of the tool(s) used in this rule. Choose from the available tool categories based on the tool's capabilities and usage pattern. This should represent the broad functional type of the tool (e.g., DATA_PROCESSING, SEARCH_ENGINE, COMPUTATION).

Return ONLY valid JSON matching this schema:
{
  "_id": 0,
  "domain": "FAMILIAL_RELATIONSHIPS",
  "qualifier": ["SEQUENTIAL", "MULTI_STEP"],
  "action": ["DECOMPOSE", "VALIDATE"],
  "strength": "MANDATORY",
  "tool_category": "DATA_PROCESSING"
}
\end{lstlisting}

\subsection{Sample Rules}
\label{app:sample_rules}

Below, we present a representative subset of learned rules. Each rule is shown in natural language along with its corresponding symbolic representation.

\lstdefinestyle{rulestyle}{
  basicstyle=\ttfamily\small,
  breaklines=true,
  breakatwhitespace=true,
  columns=fullflexible,
  frame=single,
  backgroundcolor=\color{gray!5},
frame=single,
rulecolor=\color{blue!50!black},
  captionpos=b
}
\lstset{inputencoding=utf8}

\begin{lstlisting}[style=rulestyle, caption={Sample rules learned from the ToolHop dataset}]

1. If the user query involves identifying a complex relationship (e.g., step-relative) and the available tools do not directly support the requested relationship type, then decompose the query into intermediate subtasks that progressively resolve the relationship through simpler, directly supported relationship types.
if (domain=RELATIONSHIP_RESOLUTION and qualifier=[RELATIONSHIP_CHAIN
_TRAVERSAL, INTERMEDIATE_ENTITY_IDENTIFICATION]) then (action=[DECOMPOSE_QUERY, RESOLVE_INTERMEDIATE_ENTITY]) with strength=MANDATORY

2. If a query involves determining a property of an object derived from intermediate steps (e.g., counting, analyzing, or transforming an attribute of an entity), then the decomposition must explicitly include a subtask to apply the appropriate process or transformation to the derived attribute.
if (domain=QUERY_DECOMPOSITION and qualifier=[MULTI_STEP_REASONING, DERIVATIVE_ENTITY]) then (action=[DECOMPOSE_QUERY, TRANSFORM_INPUT]) with strength=MANDATORY

3. If a query fails due to insufficient or incomplete data returned by a tool, then refine the input arguments by including optional parameters that provide additional context (e.g., variants of names, time periods, regions, or other applicable constraints) to improve data retrieval accuracy.
if (domain=TOOL_ARGUMENT_VALIDATION and qualifier=[INCOMPLETE_DATA
_REFINEMENT]) then (action=[REFINE_INPUT_ARGUMENTS, HANDLE_INCOMPLETE_DATA]) with strength=RECOMMENDED

4. If a query involves determining an attribute (e.g., date, location, event) of an entity that is related to another entity (e.g., family member), then decompose the query to first identify the related entity before attempting to determine its attribute.
if (domain=QUERY_DECOMPOSITION and qualifier=[RELATIONSHIP_CHAIN_
TRAVERSAL, INTERMEDIATE_ENTITY_IDENTIFICATION]) then (action=[DECOMPOSE_QUERY, RESOLVE_INTERMEDIATE_ENTITY]) with strength=MANDATORY

5. If the subtask requires extracting a specific subset of information (e.g., first name from a full name), then construct arguments for the tool to restrict the output to match the requested subset by enabling relevant optional parameters explicitly, while leaving unrelated parameters at their defaults.
if (domain=NAME_PROCESSING and qualifier=[NAME_COMPONENT_EXTRACTION]) then (action=[CONSTRUCT_ARGUMENTS, REFINE_INPUT_ARGUMENTS]) with strength=MANDATORY

6. If the user query involves identifying information about a specific familial relationship across multiple generations (e.g., paternal grandfather), then decompose the query such that the familial relationship is resolved directly using tools designed to retrieve genealogical data, without redundantly extracting unrelated immediate relationships.
if (domain=RELATIONSHIP_RESOLUTION and qualifier=[FAMILIAL_RELATIONSHIP, RELATIONSHIP_CHAIN_TRAVERSAL]) then (action=[DECOMPOSE_QUERY, IDENTIFY_RELATIONSHIP]) with strength=MANDATORY

7. If a query involves determining the ancestry of a person based on a specific familial relationship (e.g., maternal grandfather), then decompose the task into subtasks that sequentially query each familial link (e.g., mother, mother's father) and ensure intermediate results are used to refine subsequent subtasks.
if (domain=HIERARCHICAL_RELATIONSHIPS and qualifier=[FAMILIAL_RELATIONSHIP, RELATIONSHIP_CHAIN_TRAVERSAL, INTERMEDIATE_ENTITY_IDENTIFICATION]) then (action=[DECOMPOSE_QUERY, SEQUENCE_SUBTASKS, RESOLVE_INTERMEDIATE_ENTITY]) with strength=MANDATORY

8. If a query involves identifying an entity indirectly related to another (e.g., the founder of a political party identified in a previous step), then ensure that intermediate subtasks chain outputs logically by querying the relationship explicitly rather than assuming associations provided in the query.
if (domain=INDIRECT_ENTITY
_IDENTIFICATION and qualifier=[RELATIONSHIP_CHAIN
_TRAVERSAL, INTERMEDIATE_ENTITY_IDENTIFICATION]) then (action=[SEQUENCE_SUBTASKS, RESOLVE_INTERMEDIATE_ENTITY]) with strength=MANDATORY

9. If a query involves retrieving information about an individual's relative, the decomposition must include a subtask to explicitly identify the relative before retrieving specific information about them.
if (domain=RELATIONSHIP_RESOLUTION and qualifier=[FAMILIAL_RELATIONSHIP, RELATIONSHIP_CHAIN_TRAVERSAL]) then (action=[DECOMPOSE_QUERY, RESOLVE_INTERMEDIATE_ENTITY]) with strength=MANDATORY

10. If the subtask involves identifying a familial or social relationship between entities, then select a tool designed to query relationships, and do not select a tool designed for extracting name components or parsing names.
if (domain=RELATIONSHIP_RESOLUTION and qualifier=[FAMILIAL_RELATIONSHIP, RELATIONSHIP_CHAIN_TRAVERSAL]) then (action=[MATCH_TOOL_TO_SUBTASK]) with strength=MANDATORY


\end{lstlisting}

\begin{lstlisting}[style=rulestyle, caption={Sample rules learned from the BFCL: Live-Multiple dataset}]

1. If a reasoning process identifies a subtask requiring tool-based execution, then ensure the tool name generated in the output matches exactly with one of the available tools in the schema. Use schema validation to confirm that the generated tool name exists before finalizing the response.
if (domain=TRANSPORTATION and qualifier=[SCHEMA_VALIDATION, TOOL_SELECTION_ERROR]) then (action=[CONFIRM_SCHEMA_COMPLIANCE, SELECT_TOOL_BASED_ON_QUERY]) with strength=MANDATORY

2. IF the query involves retrieving general information about an entity, concept, or event that is not explicitly tied to recent updates or temporal relevance, THEN select a tool designed for general web searches instead of a tool specialized for retrieving recent news or updates.
if (domain=GENERAL_INFORMATION

_RETRIEVAL and qualifier=[TOOL_SELECTION_ERROR]) then (action=[SELECT_TOOL_BASED_ON_QUERY]) with strength=MANDATORY

3. If the subtask requires checking availability or stock levels for specific product attributes (e.g., sizes, stock count), then select a tool whose schema explicitly supports inventory-related operations, rather than using a tool designed for general product details.
if (domain=INVENTORY_MANAGEMENT and qualifier=[TOOL_SELECTION_ERROR]) then (action=[SELECT_TOOL_BASED_ON_QUERY]) with strength=MANDATORY

4. If the subtask involves checking the availability of a specific attribute (e.g., size, color, or stock) for a product, then select a tool whose schema explicitly includes functionality for querying availability of attributes rather than retrieving general product details.
if (domain=INVENTORY_MANAGEMENT and qualifier=[TOOL_SELECTION_ERROR]) then (action=[SELECT_TOOL_BASED_ON_QUERY]) with strength=MANDATORY

5. If a user query explicitly provides a value for a parameter, regardless of whether the parameter has a default value in the schema, prioritize the user-provided value over the default value when constructing arguments.
if (domain=DATABASE_MANAGEMENT and qualifier=[USER_QUERY_VALUE_PRIORITY, DEFAULT_VALUE_ASSIGNMENT]) then (action=[PRIORITIZE_USER_VALUE, CONSTRUCT_TOOL_ARGUMENTS]) with strength=MANDATORY

6. If the user query explicitly requests understanding, explanation, or help about a functionality, then select a tool designed to provide informational assistance rather than one intended for operational execution.
if (domain=GENERAL_INFORMATION_RETRIEVAL and qualifier=[TOOL_SELECTION_ERROR]) then (action=[SELECT_TOOL_BASED_ON_QUERY]) with strength=MANDATORY

7. If the intended action is to retrieve information based on user-specified filters or constraints, then select a tool that explicitly supports filtering or constraint-based retrieval in its schema description. Ensure the subtask aligns directly with the tool's primary function as specified in its schema.
if (domain=GENERAL_INFORMATION_RETRIEVAL and qualifier=[TOOL_SELECTION_ERROR]) then (action=[SELECT_TOOL_BASED_ON_QUERY]) with strength=MANDATORY

8. If a subtask involves identifying available options or exploring possible choices for a resource (e.g., accommodations, flights, attractions), then select a tool designed for searching or browsing those resources rather than a tool designed for finalizing or reserving them.
if (domain=GENERAL_INFORMATION_RETRIEVAL and qualifier=[TOOL_SELECTION_ERROR]) then (action=[SELECT_TOOL_BASED_ON_QUERY]) with strength=MANDATORY

9. If the user query specifies a constraint related to an entity (e.g., actor name, director name, genre, etc.), then construct the corresponding parameter value by directly extracting the entity mentioned in the query, instead of using default or placeholder values.
if (domain=GENERAL_INFORMATION_RETRIEVAL and qualifier=[REQUIRED_PARAMETER

_FORMATTING, USER_QUERY_VALUE_PRIORITY]) then (action=[CONSTRUCT_TOOL_ARGUMENTS, PRIORITIZE_USER_VALUE]) with strength=MANDATORY

10. If a schema explicitly limits valid values for an input parameter to a predefined set, then when constructing arguments, map user preferences to the closest valid value within this set, or default to a neutral option (e.g., no filtering) when the schema disallows explicit preferences.
if (domain=GENERAL_INFORMATION_RETRIEVAL and qualifier=[SCHEMA_VALIDATION, DEFAULT_VALUE_ASSIGNMENT]) then (action=[CONSTRUCT_TOOL_ARGUMENTS, ASSIGN_DEFAULT_VALUE]) with strength=MANDATORY


\end{lstlisting}

\begin{lstlisting}[style=rulestyle, caption={Sample rules learned from the BFCL: Multi-Turn-Base dataset}]

1. If a subtask involves initiating a system or process with preconditions, then identify and include all required preconditions (e.g., state validations, sequential actions) in the task decomposition, ensuring that they are executed in the correct order prior to invoking the main action. For example, if starting a process requires preconditions A, B, and C, verify and satisfy A, B, and C sequentially before initiating the process.
if (domain=PROCESS_MANAGEMENT and qualifier=[PRECONDITION_VALIDATION, TASK_SEQUENCE_VALIDATION]) then (action=[DECOMPOSE_TASK, VALIDATE_PRECONDITIONS, ENSURE_SEQUENCE]) with strength=MANDATORY

2. If a user query involves an action that requires specific prerequisites to be met (e.g., a dependency between tasks or a state requirement), decompose the query into subtasks that explicitly address the prerequisites first. Ensure each prerequisite subtask is executed and validated before proceeding with the dependent action.
if (domain=PROCESS_MANAGEMENT and qualifier=[PREREQUISITE_TASK, DEPENDENCY_MANAGEMENT, TASK_SEQUENCE_VALIDATION]) then (action=[DECOMPOSE_TASK, VALIDATE_PRECONDITIONS, EXECUTE_SUBTASK]) with strength=MANDATORY

3. If the task involves converting a monetary target between two currencies, then ensure the base currency matches the source of the monetary target and the target currency matches the destination specified in the query context. Do not reverse these roles, as it will misalign the output with the intended goal.
if (domain=CURRENCY_CONVERSION and qualifier=[CURRENCY_CONSISTENCY, PRECONDITION_VALIDATION]) then (action=[MATCH_CURRENCY, VALIDATE_PRECONDITIONS]) with strength=MANDATORY

4. If a query requires identifying entities from a complete set (e.g., all available options), then retrieve the comprehensive set explicitly using tools designed for global enumeration before attempting subtasks that depend on specific entities. Do not attempt to infer the set by piecemeal or localized retrieval from individual components.
if (domain=ENTITY_RETRIEVAL and qualifier=[GLOBAL_ENUMERATION, DEPENDENCY_MANAGEMENT]) then (action=[RETRIEVE_GLOBAL_SET, EXECUTE_SUBTASK]) with strength=MANDATORY

\end{lstlisting}

\subsection{Use of LLMs}

ChatGPT\footnote{\url{https://chatgpt.com}} is used to polish the writing of this paper.

\end{document}